\documentclass[]{article}
\pdfoutput=1
\usepackage[space]{grffile}
\usepackage{caption}
\usepackage{subcaption}
\usepackage{multirow}
\usepackage{url}
\usepackage{algorithmic}
\usepackage{algorithm2e}
\usepackage{amsmath,amsfonts,amssymb}
\usepackage{graphicx}
\usepackage{cite}
\usepackage{geometry}
\usepackage{subcaption}
\usepackage{cleveref}
\usepackage{float}
\title{Effect of Various Regularizers on Model Complexities of Neural Networks in Presence of Input Noise}
\author{Mayank Sharma, Aayush Yadav, Sumit Soman, Jayadeva }

\begin{document}
\date{}
\maketitle

\begin{abstract}
Deep neural networks are over-parameterized, which implies that the number of parameters are much larger than the number of samples used to train the network. Even in such a regime deep architectures do not overfit. This phenomenon is an active area of research and many theories have been proposed trying to understand this peculiar observation. These include the Vapnik Chervonenkis (VC) dimension bounds and Rademacher complexity bounds which show that the capacity of the network is characterized by the norm of weights rather than the number of parameters. However, the effect of input noise on these measures for shallow and deep architectures has not been studied.
In this paper, we analyze the effects of various regularization schemes on the complexity of a neural network which we characterize with the loss, $L_2$ norm of the weights, Rademacher complexities (Directly Approximately
Regularizing Complexity-DARC1), VC dimension based Low Complexity Neural Network (LCNN) when subject to varying degrees of Gaussian input noise. We show that $L_2$ regularization leads to a simpler hypothesis class and better generalization followed by DARC1 regularizer, both for shallow as well as deeper architectures. Jacobian regularizer works well for shallow architectures with high level of input noises. Spectral normalization attains highest test set accuracies both for shallow and deeper architectures. We also show that Dropout alone does not perform well in presence of input noise. Finally, we show that deeper architectures are robust to input noise as opposed to their shallow counterparts.
\end{abstract}

\section{Introduction}

Deep convolutional neural networks have recently become the de-facto approach for feature construction and classification in computer vision. With applications from image classification for object recognition \cite{krizhevsky2012imagenet, simonyan2014very, szegedy2015going, he2015delving, he2016identity} detection \cite{ren2015faster, girshick2015fast} to end-to-end learning for complex computer vision tasks \cite{antol2015vqa, karpathy2015deep}. Their impressive performance on a variety of tasks have made them immensely applicable to a variety of applications, including transfer learning \cite{nam2016learning, dosovitskiy2014discriminative}. The widespread general applicability of deep neural network architectures such as Residual Networks \cite{he2016identity}, Inception models \cite{szegedy2015going} and Wide Resnet \cite{zagoruyko2016wide} have made them popular recognition architectures for various applications.\\

The generalization properties of deep neural networks have been studied in great depth over the last few years. This has further resulted in their widespread adaptation into real world scenarios. Moreover, coupled with popular regularization methods such as Batch Normalization \cite{ioffe2015batch}, Dropout \cite{srivastava2014dropout} and $L_2$-Regularization these models have been able to achieve significant improvements in accuracies by further reducing the generalization errors.
Recent works in this area have led to new distribution dependent regularization priors such as direct regularization of model complexity \cite{kawaguchi2017generalization} and taking a bounded spectral norm of the network’s Jacobian matrix in the neighbourhood of the training sample \cite{sokolic2017robust}. While each of these methods solves a unique problem, we seek to look at the overall model in terms of the network complexity and its relative accuracies on a test set.

The goal of this study is to identify an empirically backed strong regularizer that can reduce the complexity of the network by resulting in smaller weights and perform equally well in the presence of input noise. By comparing the accuracies on a hold out dataset we also show that the effects of regularization vary greatly across these different methods. 
TO best of our knowledge this is the first such study of regularization methods in deep learning under complexity bounds and varying input noise. Our primary contribution lies in demonstrating  strong experimental evidence in favor of certain regularizers over the others. We show that deep networks are robust in presence of input noise and $L_2$ norm acts as a proxy for model complexity of neural networks. We also show that distribution dependent DARC1 regularizer and $L_2$ regularizer perform well in presence of input noise in the training set since the network tends to prefer a simpler hypothesis to get better test set accuracies on a clean noise-free test set.

The paper is divided into 5 sections. Section \ref{sec:prelim} introduces  pre-requisites to generalization error and presents a brief mathematical background of various measures of complexities and regularizers. Section \ref{sec:experiments} discusses the experimental settings used in the paper along with the results on accuracies and various notions of model complexities and techniques to control the complexities like $L_2$ norm, DARC1 norm, Dropout, Jacobian norm and Spectral regularization. Section \ref{sec:discussion} presents insights and key findings obtained from the experiments Finally Section \ref{sec:conclusion} presents conclusions and future line of research.

\section{Preliminaries}
\label{sec:prelim}
In this section, we describe the background, measures of complexities and the regularizers that control the model complexity of the neural network. Firstly, we present various notations used in the paper in the Table \ref{tab:notations}. 
\begin{table}[hbtp]
	\centering
	\caption{Notations}
	\scalebox{0.8}{
		\begin{tabular}{|c|c|c|}
			\hline
			{S.no.} & Notation & Description  \\
			\hline
			1     & $\mathcal{X}$ & Input space \\
			2     & $\mathcal{Y}$ & Label space \\			
			3     & $x \in \mathcal{X}$ & Input sampled from $\mathcal{X}$\\
			4     & $y \in \mathcal{Y}$ & Output sampled from $\mathcal{Y}$\\
			5     & $P$ & Distribution of $(x,y)$ \\
			6     & $F$ & Set of functions/hypothesis space\\
			7     & $\ell$ & Loss function\\
			8 	  & $\mathcal{L}_{F}$ & Family of loss functions associated with $F$.\\
			9     & $m$ & \# training samples \\
			10    & $d$ & \# input dimension \\
			11    & $K$ & \# classes \\
			12    & $R[f]$ & $\mathbb{E}_{x,y \sim P}[\ell(f(x),y)]$, expected risk of a function $f \in F$\\
			13    & $f_{\mathcal{A}(S_{m})}: \mathcal{X} \rightarrow \mathcal{Y}$ & Model learned by learning algorithm $\mathcal{A}$\\
			14    & $S_{m}$ & i.i.d. training dataset $\{(x_1,y_1) \ldots (x_m,y_m)\}$ of size $m$\\
			15    & $\hat{R}_{S_{m}}[f]$ & $\frac{1}{m}\sum_{i=1}^{m}\ell(f(x_i),y_i)$, empirical risk of $f$\\
			16    & $\eta$ & Noise level \\
			17    & $\lambda$ & Weight decay/DARC1/Jacobian/LCNN norm hyperparameter \\
			18 & $\gamma$ & margin of a classifier \\
			20 & $VCdim(\mathcal{F})$  & Vapnik-Chervonenkis dimension of the hypothesis class $\mathcal{F}$ \\
			\hline
		\end{tabular}%
	}
	\label{tab:notations}%
\end{table}%

The goal in machine learning is to minimize the expected risk $R[f_{\mathcal{A}(S_{m})}]$. However, the expected risk being non-computable, we aim to minimize the computable empirical risk denoted by $\hat{R}_{S_{m}}[f_{\mathcal{A}(S_{m})}]$. The generalization gap is given by:

\begin{gather}
R[f_{\mathcal{A}(S_{m})}] - \hat{R}_{S_{m}}[f_{\mathcal{A}(S_{m})}]
\end{gather}

A major drawback of this approach is the dependence of $f_{\mathcal{A}(S_{m})}$ on the same dataset $S_{m}$ used in the definition of $\hat{R}_{S_{m}}$. One way to tackle this is to considering the worst-case gap for functions in the hypothesis space.

\begin{gather}
R[f_{\mathcal{A}(S_{m})}] - \hat{R}_{S_{m}}[f_{\mathcal{A}(S_{m})}] \leq \sup_{f \in F}R[f] - \hat{R}_{S_{m}}[f] 
\end{gather}

A union bound over all the elements of the hypothesis class yields a vacuous bound, hence we consider other quantities to characterize the complexity of $F$ namely, Vapnik-Chervonenkis (VC) dimension \cite{bartlett1997valid,sontag1998vc} and Rademacher Complexity \cite{bartlett2002rademacher,bartlett2017spectrally,neyshabur2015norm}.
If the codomain of the loss is given by $\ell \in [0,1]$, for any $\delta > 0$ with probability at least $1-\delta$,

\begin{gather}
\sup_{f \in F}R[f] - \hat{R}_{S_{m}}[f] \leq 2\mathfrak{R}_{m}(\mathcal{L}_{F}) + \sqrt{\frac{\ln\frac{1}{\delta}}{2m}}
\end{gather}

where, $\mathfrak{R}_{m}(\mathcal{L}_{F})$ is the Rademacher Complexity of $\mathcal{L}_{F}$, which can then be bounded by Rademacher complexity of F denoted as $\mathfrak{R}_{m}(F)$. Similar bounds can be found in the literature using VC dimension, fat shattering dimension and covering numbers. We now highlight the various bounds for a neural network in the next subsection.

\newpage
\subsection{Measures of Complexities of a Neural Network}

Consider a deep net with $H$ layers $A^1,A^2, \ldots A^H$ and output margin $\gamma$ on training set $S_m$. Various generalization bounds proposed in the literature are mentioned in the Table \ref{tab:bounds}. The term $\rho$ denotes the number of parameters of a multilayered feed forward neural network.

\begin{table}[hbtp]
	\centering
	\caption{Notations}
	\scalebox{0.8}{
		\begin{tabular}{|c|c|c|}
			\hline
			{S.no.} & Reference & Measure  \\
			\hline
			1   & Sontag \cite{sontag1998vc} & $\mathcal{O}(\rho \log\rho)$ \\
			2   & Bartlett et al. \cite{bartlett2002rademacher} & $\frac{1}{\gamma^2}\prod_{i=1}^{H}\|A^i\|_{1,\infty}$\\
			3   & Neyshabur et al. Sharma et. al. \cite{neyshabur2015norm,sharma2018radius} & $\frac{1}{\gamma^2}\prod_{i=1}^{H}\|A^i\|_{F}^2$\\
			4   & Bartlett et al. \cite{bartlett2017spectrally} & $\frac{1}{\gamma^2}\prod_{i=1}^{H}\|A^i\|_2^2\sum_{i=1}^{H}\frac{\|A^i\|_{1,2}^2}{\|A^i\|_{2}^2}$\\
			5   & Neyshabur et al. \cite{neyshabur2017exploring,neyshabur2017pac} & $\frac{1}{\gamma^2}\prod_{i=1}^{H}\|A^i\|_2^2\sum_{i=1}^{H}h_i\frac{\|A^i\|_{F}^2}{\|A^i\|_{2}^2}$	\\	
			6   & Kawaguchi et al. \cite{kawaguchi2017generalization} & $\frac{\lambda}{m}\left(\max_k\sum_{i=1}^{m}|f_{k}^{H+1}(x_i)|\right)$	\\		
			7   & Pant et al. \cite{kawaguchi2017generalization} & $\frac{\lambda}{m}\left(\sum_{i=1}^{m}\left(f_{k}^{H+1}(x_i)\right)^2\right)$	\\		
			
			\hline
		\end{tabular}%
	}
	\label{tab:bounds}%
\end{table}%

The expression $\sum_{i=1}^{d}h_i\frac{\|A^i\|_{F}^2}{\|A^i\|_{2}^2}$ is the sum of stable ranks of the layers which is a measure of the parameter count. The expression $\prod_{i=1}^{d}\|A^i\|_2^2$ is related to the Lipschitz constant of the network. However, for this paper we use $\sum_{i=1}^{d}\|A^i\|_{F}^2$ as a measure of the network complexity.

\subsection{Regularizers}
We now briefly describe various regularizers used in neural networks namely, $L_2$ norm, dropout, jacobian, DARC1 and spectral normalization. We study the behavior of these regularizers in presence of varying input noise. In doing so, we characterize the complexity of network by the $L_2$ norm of the parameters and DARC1 Rademacher bound. 

\begin{itemize}
	\item \textbf{$L_2$ norm}: Weight decay or $L_2$ regularization is one of the most popular regularization techniques used to control the model complexity. It amounts to penalizing the $L_2$ norm of the weights of the network. Almost all the generalization bounds are a function of $L_2$ norm of the weights. Sontag \cite{sontag1998vc} first proposed the VC dimension of a multilayer neural network in terms of the number of parameters. But for a network with millions of parameters, the bound turns out to be vacuous as the network seems to generalize well with much smaller samples than the number of parameters. Bartlett \cite{bartlett1997valid} also proposed the VC dimension or fat shattering dimension of a multilayer feed forward neural network in terms of the product of $L_\infty$ norm of the weights. This bound showed that in order to control the complexity of a network, one has to regularize the norm of the weights. Neyshabur \cite{neyshabur2015norm} presented the generalized analysis of complexity in terms of $L_{p,q}$ where $\frac{1}{p} + \frac{1}{q} = 1$. Some other recent bounds are mentioned in table \ref{tab:bounds}.
	
	\item \textbf{DARC1 norm}: Kawaguchi et al. \cite{kawaguchi2017generalization} suggested minimizing the max norm of the activation as a regularizer. They termed the method as Directly Approximately Regularizing Complexity (DARC) and named a basic version of their proposed regularization prior as DARC1. They argue that the common generalization bounds (as mentioned in the table) are too loose to be used practically. They therefore consider a margin based 0-1 loss defined as:
	\begin{equation}
		\ell_{\gamma,m}(z)=\begin{cases}
		0, & \text{if $z > \gamma$}.\\
		1 - \frac{z}{\gamma}, & \text{if $ 0 < z \leq \gamma$}.\\
		1, & \text{if $z \leq 0$}.\\
	\end{cases}
	\end{equation}
	Finally, they show that 
	\begin{gather}
		R[f] \leq \hat{R}_{m,\gamma}[f] + \frac{K^2}{\gamma m}\hat{\mathfrak{R}}_m(F) + \sqrt{\frac{\ln \frac{1}{\delta}}{2m}}
	\end{gather} 
	where, $\hat{\mathfrak{R}}_m(F)$ is the Rademacher complexity of the hypothesis class $F$ defined as $\mathbb{E}_{S_m,\xi}\left[\sup_{k,f_{k}^{H+1}}\sum_{i=1}^{m}\xi_i f_{k}^{H+1}(x_i)\right]$, where $\gamma$ is the margin, $\xi_i$ are the Rademacher variables, supremum is taken over all $k \in \{1,\ldots, K\}$ and $f_k^{H+1}$ allowed in $F$. $\delta$ is the confidence level and $y(x) \in \{1,\ldots,K\}$ is the true label of $x$.
	Instead of using worst case vacuous bounds, they use the approximation of $\hat{\mathfrak{R}}_m(F)$ with an expectation over the known dataset $S_m$. DARC1 is the new regularization term added on each minibatch as follows:
	\begin{gather}
		\text{loss} = \text{original loss} + \frac{\lambda}{m}\left(\max_k\sum_{i=1}^{m}|f_{k}^{H+1}(x_i)|\right)
	\end{gather}
	Following Zhang et al. \cite{zhang2016understanding}, the regularizer can be seen as penalizing the most confident predictions, thereby reducing the tendency of the model to overfit. A similar analysis is done using Low complexity Neural Network (LCNN) loss in Jayadeva et al. \cite{pant2017learning}, which instead of using a max-norm, uses $L_2$ norm over the hypothesis class of the final layer of the network. Here, original loss can be any of the standard loss function used in the literation \textit{viz}. cross-entropy, max-margin, 0-1 etc.
	
	\item \textbf{LCNN norm}: Low Complexity Neural Network \cite{pant2017learning} regularizer tries to upper bound the VC dimension of neural network using radius margin bound. It is known that the VC dimension $\gamma$ of a large margin linear classifier is upper bounded by:
	\begin{gather}
		VCdim \leq 1 + \min\left(\frac{r^2}{\gamma^2},d\right)
	\end{gather}
	where $r$ is the radius of the data, $\gamma$ is the margin and $d$ is the number of input features.
	A similar analysis can be performed for the last layer of a neural network. Here, we state without proof that the VC dimension ($VCdim$) of a neural network is bounded by:
	
	\begin{gather}
	VCdim \leq \sum_{i=1}^{m}\left(f_{k}^{H+1}(x_i)\right)^2
	\end{gather}
	
	LCNN regularization term is added on each minibatch as follows:
	\begin{gather}
	\text{loss} = \text{original loss} + \frac{\lambda}{m}\left(\sum_{i=1}^{m}\left(f_{k}^{H+1}(x_i)\right)^2\right)
	\end{gather}
	
	LCNN term penalizes the large values in the last layer and acts as a confidence penalty. Here, original loss can be any of the standard loss function used in the literation \textit{viz}. cross-entropy, max-margin, 0-1 etc.
	
	\item \textbf{Jacobian Regularizer}: Sokolic et al. \cite{sokolic2017robust} have argued that the existing generalization bounds in deep neural networks (DNN) grows disproportionately to the number of training samples. To resolve this, they propose a new lower bound expressed as a function of the network's Jacobian matrix which is based on the robustness framework of \cite{xu2012robustness}. 
	The Jacobian matrix of a DNN is given by:
	\begin{gather}
	J(x) = \frac{\partial f^{H+1}(x)}{\partial x} = \prod_{l=1}^{H+1}\frac{\partial f^{l}(z^{l-1})}{\partial z^{l-1}} . \frac{\partial f^{1}(x)}{\partial x}
	\end{gather} 
	
	Addition of Jacobian regularizer also allows the network to become robust to changes in the input. It has the effect of inducing a large classification margin at the input.
	Following theorem 4 in \cite{sokolic2017robust} the classification margin for a point $s_i = (x_i,y_i)$ with score $o(s_i) = \min_{j \neq y_i}\sqrt{2}\left(\delta_{y_i} - \delta_j\right)^T f^{H+1}(x_i)$is given by:
	\begin{gather}
		\gamma \geq \frac{o(s_i)}{\sup_{x:\|x-x_i\|_2 \leq \gamma}\|J(x)\|_2}
	\end{gather} 
	here $\delta_i$ is the Kronecker delta vector with $\delta_{ii} = 1$.
	Hence, the generalization error is bounded by:
	\begin{gather}
		R[f] \leq \hat{R}_{m,\gamma}[f] + \sqrt{\frac{2^{k+1}.(C_M)^k}{\gamma^k m}}
	\end{gather}
	Here, $C_M$ is a constant defining the $k$ dimensional manifold. It can be seen that $\|J(x)\|_2 \leq \prod_{i=1}^{H} \|A^i\|_2 $. The above bound shows that the generalization error does not increase with the number of layers provided the spectral norm of the weight matrices are bounded. If we assume the weight matrices contains orthonormal rows then the generalization error depends on the complexity of data manifold and not on depth.
	The Jacobian regularizer is given by:
	\begin{gather}
		\text{Jacobian regularizer} = \frac{1}{m}\sum_{i=1}^{m}\|J(x)\|_F^2
	\end{gather}
	
	\item \textbf{Spectral Normalization}: Spectral normalization controls the Lipschitz constant of the network by constraining the spectral norm $(\sigma(.))$ of each layer $f^i : h_{in} \rightarrow h_{out}$. The Lipschitz constant of a general differentiable function is the maximum singular value of its gradient over its domain.
	\begin{gather}
		\|f\|_{Lip}  = \sup_{x}\sigma(\nabla f(x))
	\end{gather}
	For composite functions, $\|g \circ f\|_{Lip} \leq \|g\|_{Lip}\|f\|_{Lip}$. Spectral normalization proposed for generative adversarial net \cite{miyato2018spectral}, replaces each weight $W$ with $\frac{W}{\sigma(W)}$. The computation is done using power iteration method. Let us consider a linear map $W : \Re^n \rightarrow \Re^m $. Let $v \in \Re^m$ be a vector in the domain of matrix $W$ and $u \in \Re^n$ be a vector in the codomain. Power iteration involves the following recurrence relation.
	\begin{gather}
		v_{t+1} = \frac{W^TWv_t}{\|W^TWv_t\|}
	\end{gather}
	On further simplification (see Algorithm 1 in  \cite{miyato2018spectral}), we have the relation:
	\begin{gather}
		v_{t+1} = W^Tu_{t+1}/\|W^Tu_{t}\|_2\\
		u_{t+1} = W v_{t+1}/\|W v_{t+1}\|_2 
	\end{gather}
	Finally, the weight matrix $W$ is updated as:
	\begin{gather}
		W = W/\sigma(W)
	\end{gather}
	where, $\sigma(W) = u^TWv$
	
	Authors in \cite{miyato2018spectral} argue that the gradient regularizer proposed in \cite{gulrajani2017improved} which is similar in concept to the Jacobian regularizer proposed in  \cite{sokolic2017robust}, has a drawback that the Jacobian regularizer is not able to regularize the function at the points outside of the support of the current distribution. They also show that spectral normalization does not get destabilized by large learning rates, whereas Jacobian regularizer falters with aggressive learning rates. 
	
	Spectral normalization has another advantage in terms of controlling the model complexity. Following the Bound 5 of Neyshabur et. al. \cite{neyshabur2017exploring,pant2017learning} presented in the Table \ref{tab:bounds}  as $\frac{1}{\gamma^2}\prod_{i=1}^{H}\|A^i\|_2^2\sum_{i=1}^{H}h_i\frac{\|A^i\|_{F}^2}{\|A^i\|_{2}^2}$, setting the term $\|A^i\|_{2}$ close to 1, we get the bound
	\begin{gather}
		\frac{1}{\gamma^2}\sum_{i=1}^{H}h_i\|A^i\|_{F}^2 \label{eq: spec_bound}
	\end{gather}

	The bound in eq \ref{eq: spec_bound} shows that, in order to control to complexity of the model, one needs to perform spectral normalization and $L_2$ normalization. In this case only, sum of norm of weights of the network truly indicate the capacity of the architecture.
	
	\item \textbf{Dropout}: It is known that models with large number of parameters such as deep neural architectures can model very complex functions and phenomenon. This also means that such models have a tendency to overfit. Thus regularizing such a model is imperative for good performance on the unseen test set. Dropout is such a technique proposed by Srivastava et al.  \cite{srivastava2014dropout}. Dropout works by randomly and temporarily deleting neurons in the hidden layer during the training with probability $p$. During testing/prediction, we feed the input to unmodified layer, but scale the layer output by $\frac{1}{p}$. Dropout acts as an averaging scheme on the output of a large number of networks, which is often found to be a powerful way of reducing overfitting.
	
	Also, since a neuron cannot rely on the presence of other neurons, it is forced to learn features that are not dependent on the presence of other neurons. Thus the network learns robust features, and are less susceptible to noise. This reduces the co-adaptation of features.
	
\end{itemize}
The following section describes the experimental settings and results by varying the input noise.

\newpage
\section{Experiments and Results}
\label{sec:experiments}
In this paper we analyze various regularizers and their effect on input noise. We characterize the complexity required by each network with different regularizers as the noise in the input increases. Following Arora et al. \cite{arora2018stronger}, we find that deep nets are noise-stable. However, shallow nets are not as is evident from the results.

\subsection{Experimental settings}
We describe the experimental settings used in the paper. To begin with, we add a class specific gaussian noise to each input example in the training set while the validation and test set remained unchanged. 

Let $\eta_i$ be the standard deviation of the training points belonging to class i. To each sample of class i in training set, we add a zero mean unit variance Gaussian noise scaled by $\eta_i$ and a noise level. The noise level ranges from 0 to 1.2 in steps of 0.2. The procedure to generate noisy datasets is described in the Algorithm \ref{algo:noisydataset}.

\begin{algorithm}[H]
	\DontPrintSemicolon 
	\KwIn{Training set $S_m = \{(x_1,y_1) \ldots (x_m,y_m)\} = \{X,Y\}$, $t = $ noise factor}
	\KwOut{Noisy set $\hat{S}_m$}
	$K = \#$ Classes\\
	$\hat{S}_m = []$ \\
	\For{$j := 0$ \textbf{to} $K$} {
		$temp  = X[Y == j,]$\\
		$Y1  = Y[Y == j,]$\\
		$M = \#$ samples of class j \\ 
		$\eta_j = std(temp,1)$ \\
		$I = diag(ones(N))$ \\
		$X1 = t \eta_j \mathcal{N}(0,I) + temp$ \\
		$\hat{S}_m  = [\hat{S}_m, (X1,Y1)]$
	}
	\Return{$\hat{S}_m$}
	\caption{Creating noise dataset $\hat{S}_m$}
	\label{algo:noisydataset}
\end{algorithm}
We include two types of datasets in this work. Four non-image datasets and one image dataset. We trained two networks, a shallow network for non-image datasets and a deeper network for image dataset. The details of the networks are presented below,
\begin{itemize}
	\item A network with two convolutional layers of filter size 30 and 20 respectively, and a fully connected layer of size $\left \lfloor{\frac{d - 1}{2}}\right \rfloor $, trained for 50 epochs with ADAM optimizer \cite{kingma2014adam}.
	\item A Wide Resnet 28-10 trained for 50 epochs, also with ADAM optimizer.
	\item Both these networks have Batch normalization applied on each layer.
	\item The initial learning rate is set to $10^{-3}$ for both the networks.
\end{itemize}

The shallow network was trained separately with the seven different regularizers \textit{viz.} no regularization, $L_2$ norm, DARC1 norm, LCNN norm, Dropout, Jacobian and Spectral normalization. The deep network was trained with the aforementioned regularizers. 
We tuned each network on the validation set and report the results on the test set for the best hyperparameter setting.

For the shallow conv net we used the hyperparameter settings mentioned in the Table \ref{tab:hyperparameters_shallow}.
\begin{table}[H]
	\centering
	\caption{Notations}
	\scalebox{0.8}{
		\begin{tabular}{|c|c|c|}
			\hline
			{S.no.} & Hyperparameter & Range  \\
			\hline
			1     & $L_2$ weight decay & $[10^{-4},10^{-3},10^{-2},10^{-1}]$\\
			2     & DARC1 &$[10^{-4},10^{-3},10^{-2},10^{-1}]$\\
			3     & Jacobian regularizer & $[10^{-4},10^{-3},10^{-2},10^{-1}]$\\
			4     & Dropout & $[0.2,0.4,0.6,0.8]$\\
			5     & LCNN & $[10^{-4},10^{-3},10^{-2},10^{-1}]$\\
			6     & Batch size & 128 \\
			\hline
		\end{tabular}%
	}
	\label{tab:hyperparameters_shallow}%
\end{table}%

For the deep network, we used the hyperparameters mentioned in the Table \ref{tab:hyperparameters_deep}:
\begin{table}[H]
	\centering
	\caption{Notations}
	\scalebox{0.8}{
		\begin{tabular}{|c|c|c|}
			\hline
			{S.no.} & Hyperparameter & Values  \\
			\hline
			1     & $L_2$ weight decay & $ 10^{-6}$\\
			2     & DARC1 & $10^{-3}$\\
			2     & Jacobian regularizer & $10^{-3}$\\
			2     & LCNN & $10^{-3}$\\
			3     & Dropout & $0.5$\\
			4     & Batch size & 32 \\
			\hline
		\end{tabular}%
	}
	\label{tab:hyperparameters_deep}%
\end{table}%

\subsection{Datasets}
We show results on 5 datasets in this paper. We experimented with two neural network architectures, one shallow convolutional net (with three layers) for datasets 1 to 4 and one deep (Wide Resnet 28-10) for dataset 5 (CIFAR 10). The datasets are described in the Table \ref{tab:datasets}.
\begin{table}[hbtp]
	\centering
	\caption{Datasets used in the paper}
	\scalebox{0.8}{
		\begin{tabular}{|c|l|c|c|c|c|c|}
			\hline
			\multicolumn{1}{|c|}{S.no.} & Dataset\_name & \multicolumn{1}{|c|}{\# Train} & \multicolumn{1}{|c|}{\# Val} & \multicolumn{1}{|c|}{\# Test} & \multicolumn{1}{|c|}{\# Features} & \multicolumn{1}{|c|}{\# Classes} \\
			\hline
			1     & Adult & 29304 & 9769  & 9769  & 14    & 2 \\
			2     & MNIST & 50000 & 5000  & 5000  & 784    & 10 \\		
			3    & Codrna & 293139 & 97713 & 97713 & 8     & 2 \\
			4    & Covertype & 348603 & 116204 & 116205 & 54    & 7 \\			
			5    & CIFAR10 & 50000 & 5000  & 5000  & 3072    & 10 \\
			\hline
		\end{tabular}%
	}
	\label{tab:datasets}%
\end{table}%

\subsection{Results}
We compare test set accuracies, log2 $L_2$ norm, DARC1 norm and loss for each datasets. Firstly, we present the results for shallow conv net and thereafter show the results for a Wide Resnet 28-10 on CIFAR10 dataset.

\begin{figure}[H]
	\begin{subfigure}[hbtp]{0.5\textwidth}
		\includegraphics[width=\textwidth]{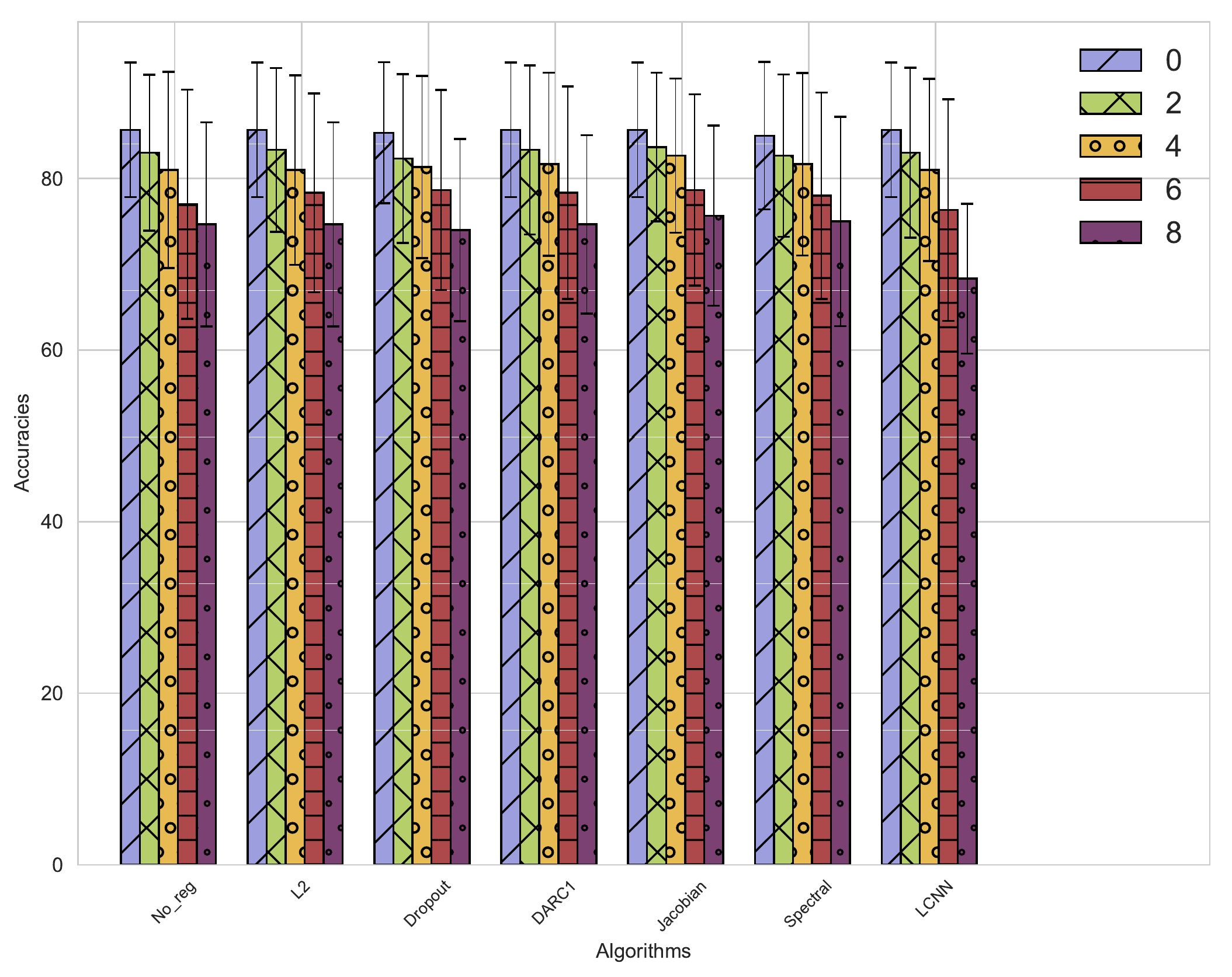}
		\caption{}
		\label{fig:acc small}	
	\end{subfigure}\qquad
	\begin{subfigure}[hbtp]{0.5\textwidth}
		\includegraphics[width=\textwidth]{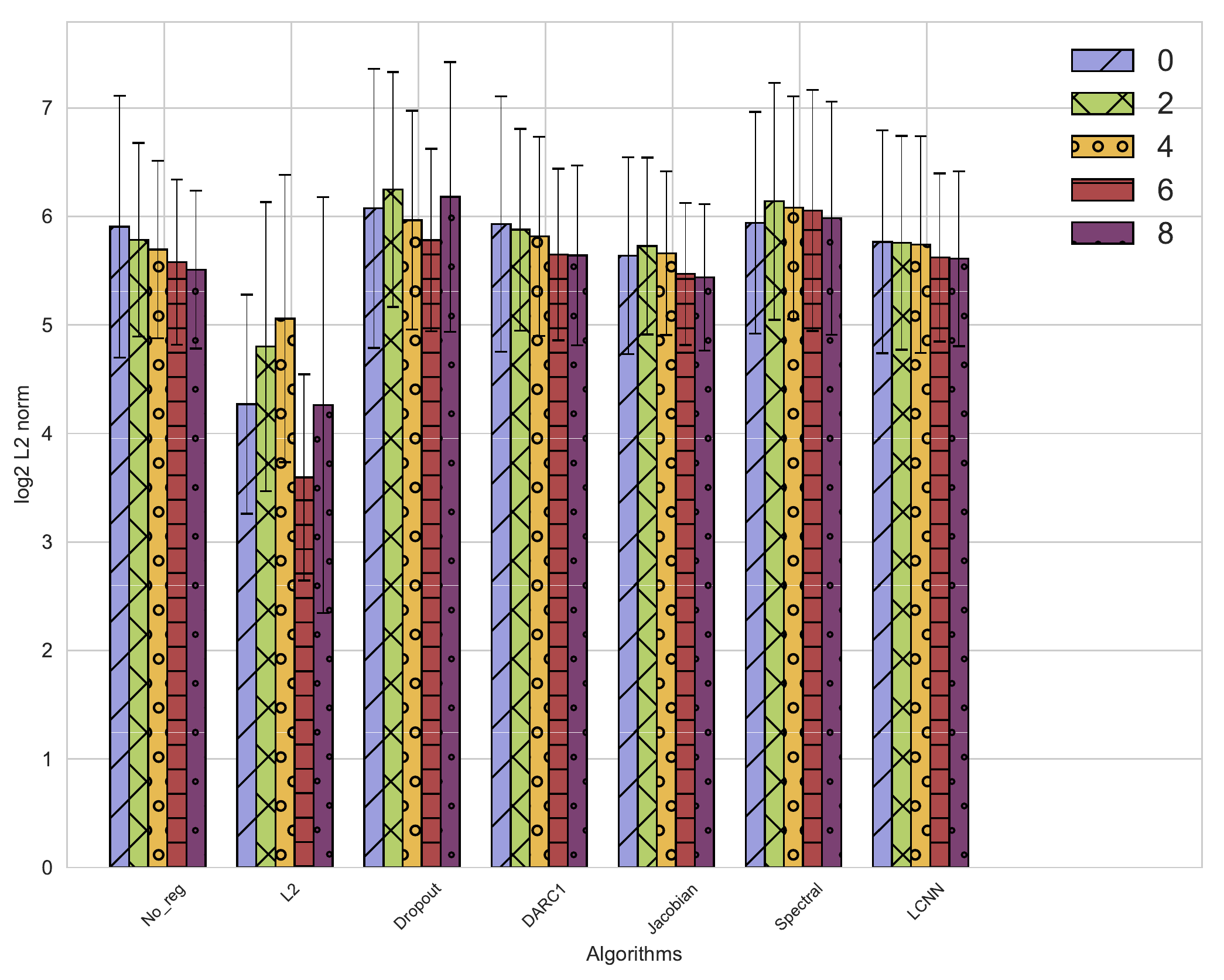}
		\caption{}
		\label{fig:log l2 small}	
	\end{subfigure}

	\begin{subfigure}[hbtp]{0.5\textwidth}
		\includegraphics[width=\textwidth]{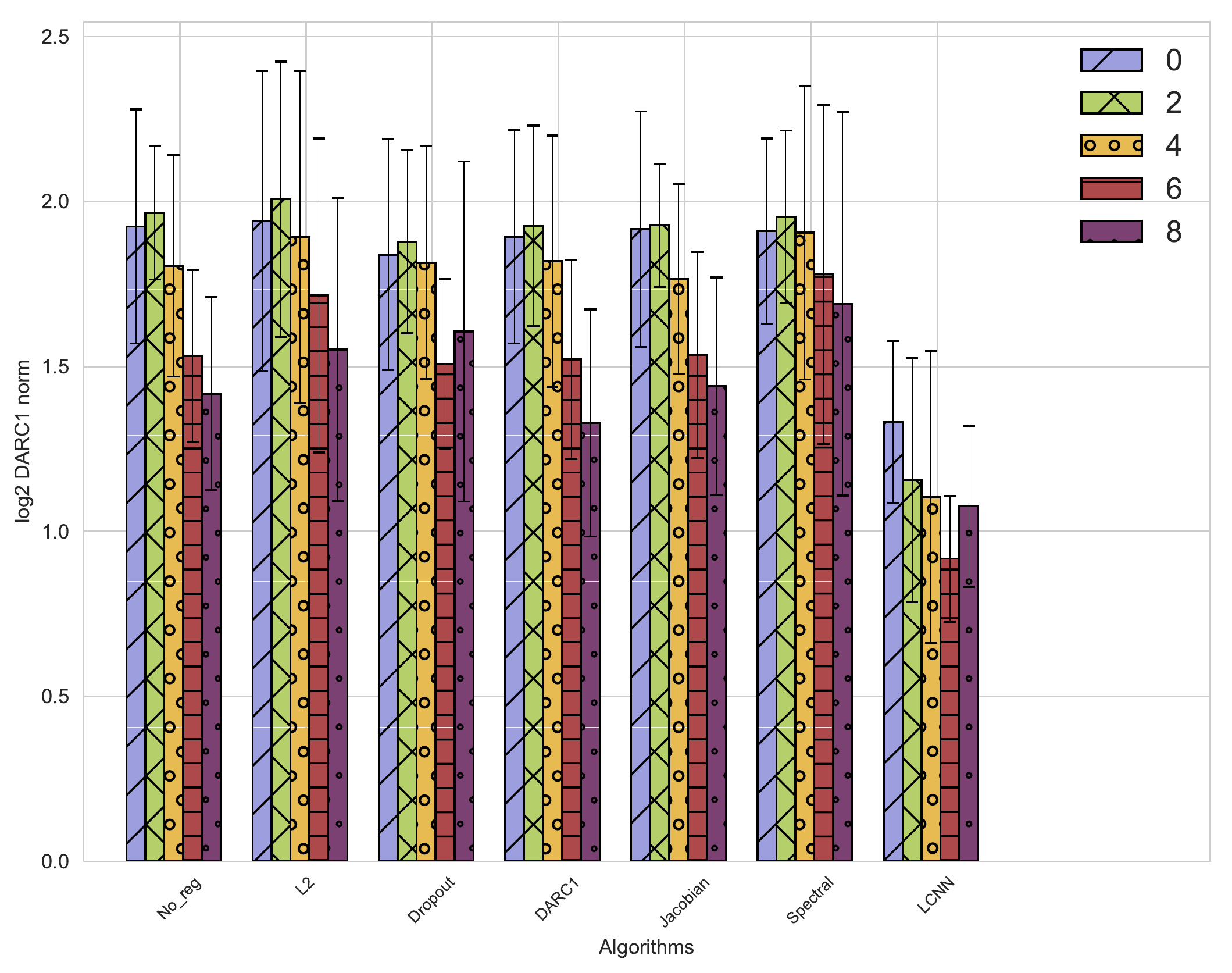}
		\caption{}
		\label{fig:log darc1 small}	
	\end{subfigure}\qquad
	\begin{subfigure}[hbtp]{0.5\textwidth}
		\includegraphics[width=\textwidth]{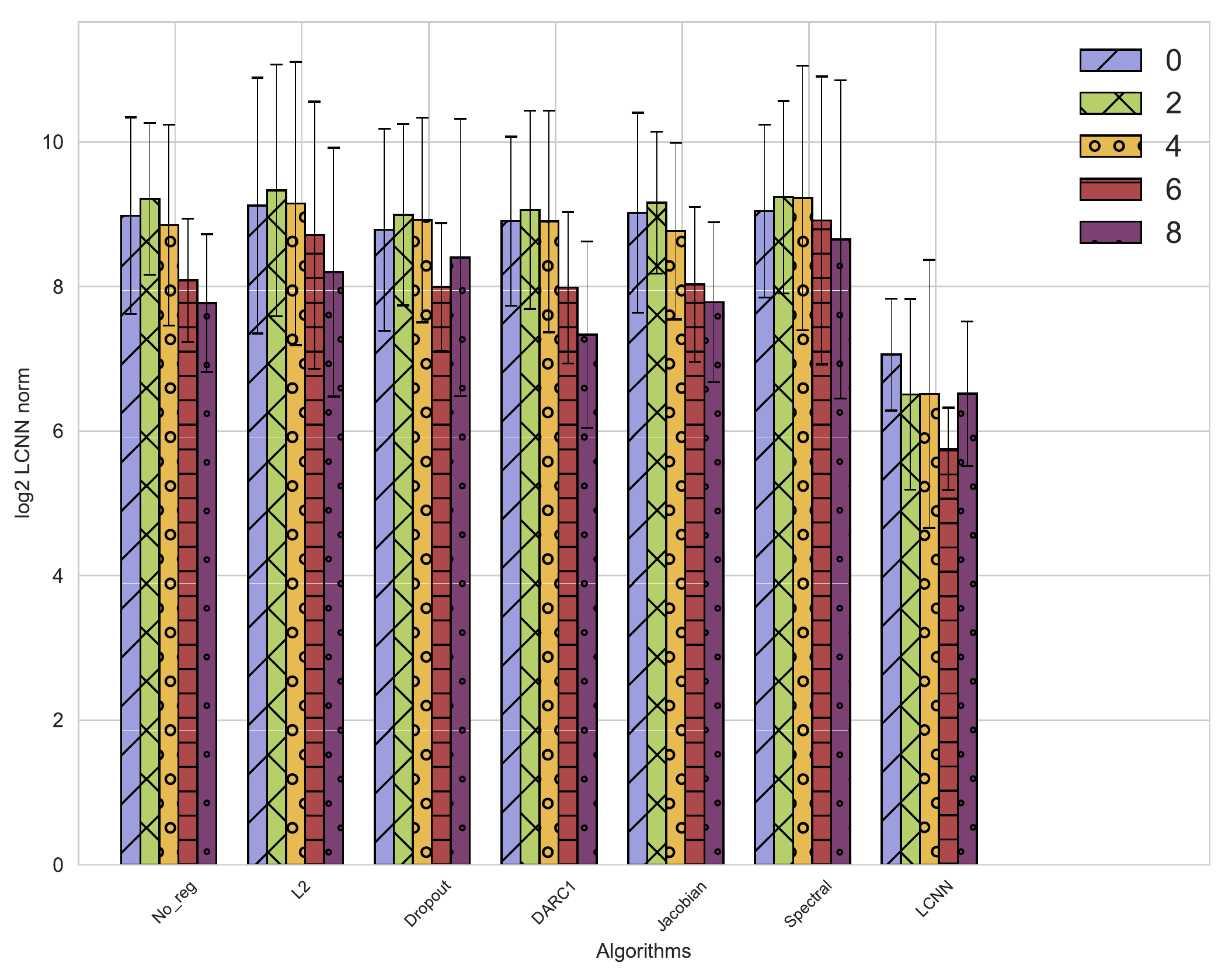}
		\caption{}
		\label{fig:log lcnn small}	
	\end{subfigure}

	\begin{subfigure}[hbtp]{0.5\textwidth}
		\includegraphics[width=\textwidth]{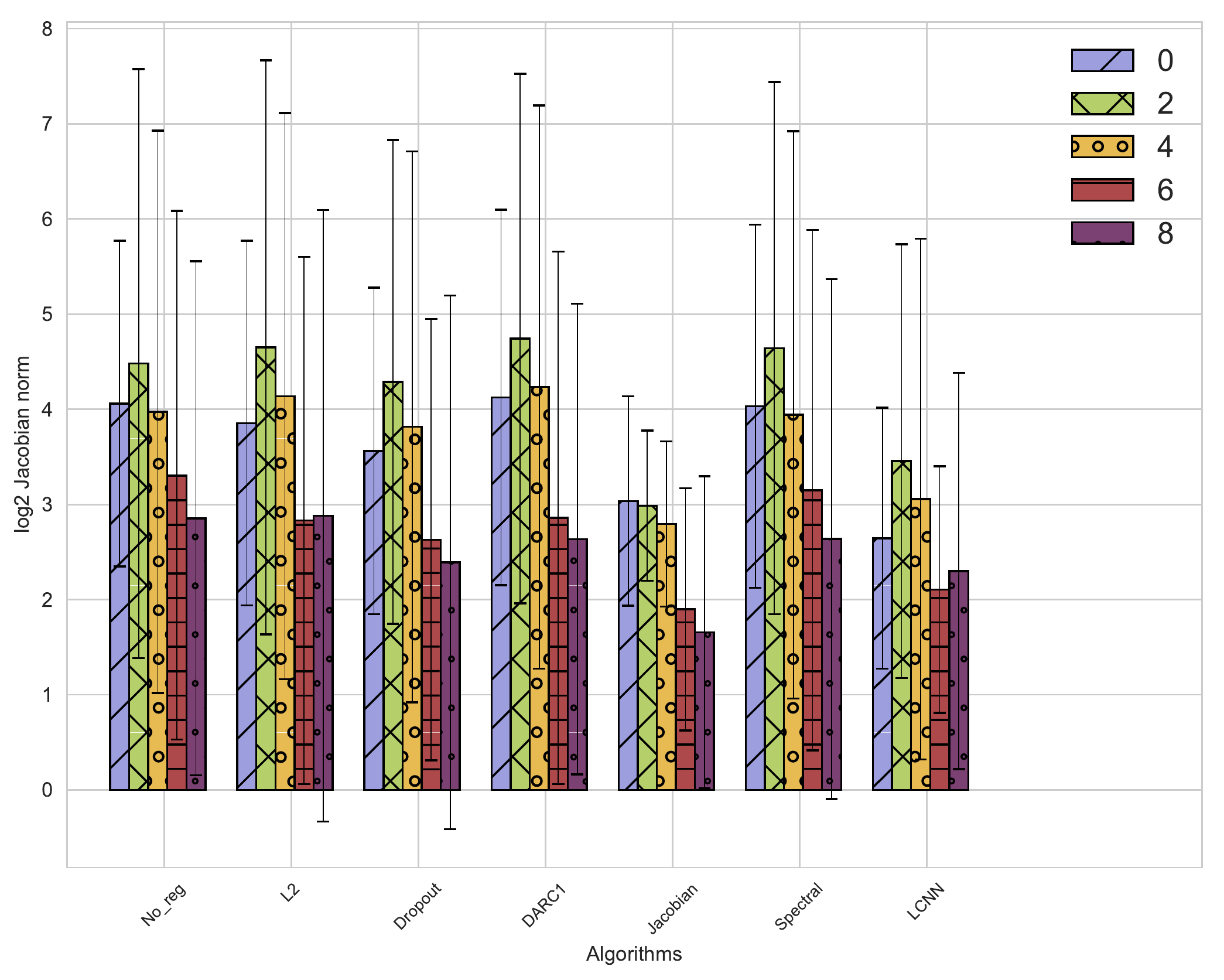}
		\caption{}
		\label{fig:log jacobian small}	
	\end{subfigure}\qquad
	\begin{subfigure}[hbtp]{0.5\textwidth}
		\includegraphics[width=\textwidth]{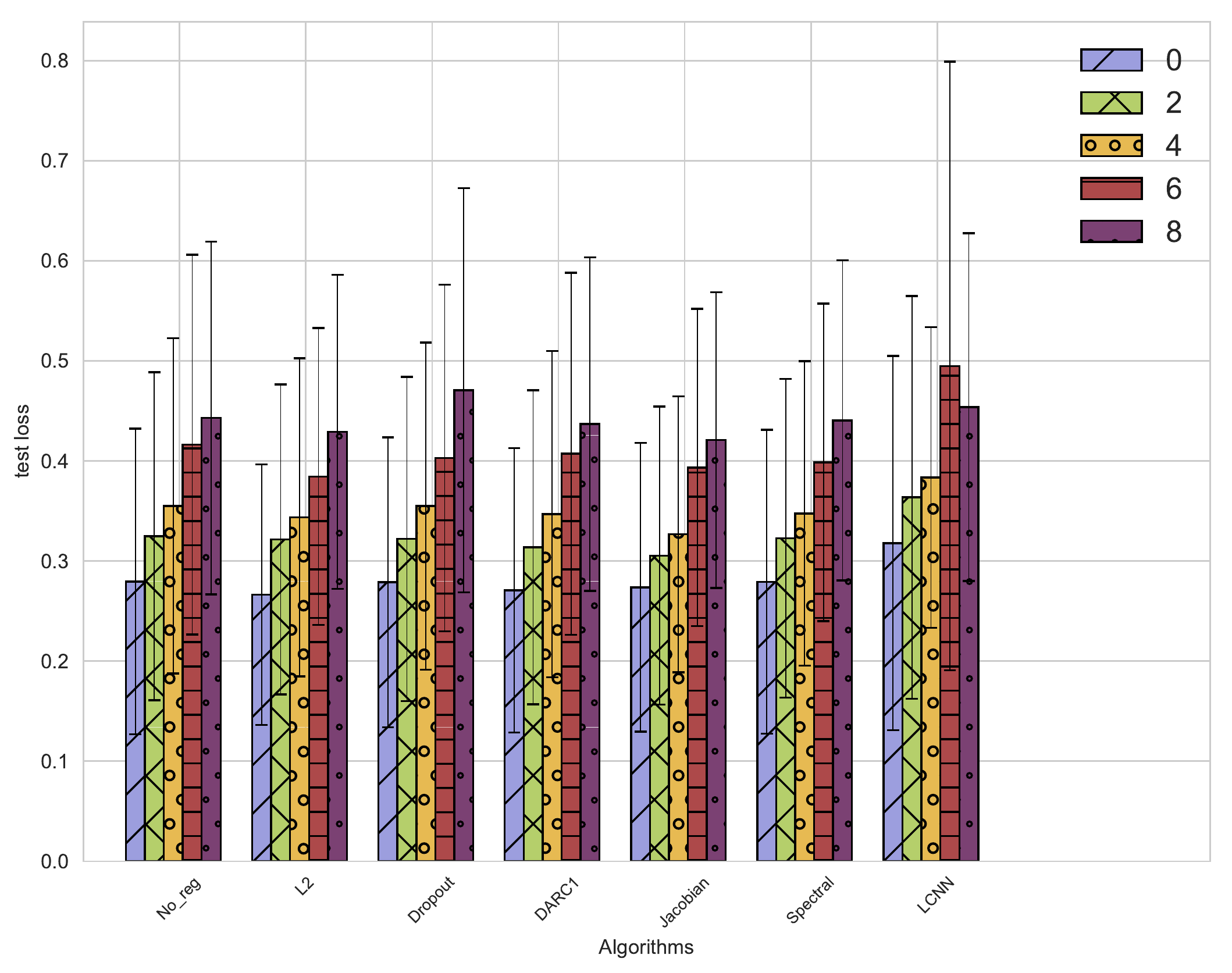}
		\caption{}
		\label{fig:loss small}	
	\end{subfigure}

	\caption{Comparison of \subref{fig:acc small} test set accuracies, \subref{fig:log l2 small} $L_2$ norm of weights, \subref{fig:log darc1 small} mean DARC1 regularizer, \subref{fig:log lcnn small} mean LCNN regularizer, \subref{fig:log jacobian small} mean Jacobian regularizer, \subref{fig:loss small} mean cross entropy loss for shallow network across datasets}
	\label{fig:comparison small}
\end{figure}

Figure \ref{fig:acc small} shows the bar plot of the mean accuracy comparison across the four datasets used for testing shallow architecture. We observe that the accuracies fall as the noise in the training set increases. Upto the noise levels of 6 there is no differences in the accuracies across regularizers, however at noise levels of 8, we see that Jacobian regularizer outperforms the others, closely followed by spectral norm regularizer and $L_2$ regularizer. LCNN regularizer however does not perform well at higher levels of noises for shallow architectures.

Figure \ref{fig:log l2 small} shows the mean log2 scaled $L_2$ norms obtained
for the shallow architecture. We observe that the network with $L_2$ regularization results in smallest $L_2$ norm. For network without regularization, DARC1, LCNN regularization we find that the $L_2$ norm decreases as the noise increases, however for Dropout, Jacobian, Spectral and $L_2$ regularization we observe an increase in $L_2$ norm initially and then a decline. The latter trend is more pronounced in case of $L_2$ norm. This is indicative of the fact that with slight increase in noise, the model increases its complexity at first to fir the noise, but as the noise increases to a larger extent, the model reduces its complexity to fit the non-noisy validation set.
Similar trends are observed for DARC1 regularizer (fig. \ref{fig:log darc1 small}), LCNN regularizer (fig. \ref{fig:log lcnn small}) and Jacobian regularizer (fig. \ref{fig:log jacobian small}). Since, LCNN is an upper bound on DARC1, the graphs have a similar trend.

Figure \ref{fig:loss small} shows the test set cross entropy loss as the noise increases. We observe an increasing trend in all the cases as noise increases.  

We now show the results of these regularizers on a deeper architecture (Wide Resnet 28-10) trained on an image dataset. For this experiment we varied the noise levels from 0 to 8 in steps of 2.

\begin{figure}[H]
	\begin{subfigure}[hbtp]{0.5\textwidth}
		\includegraphics[width=\textwidth]{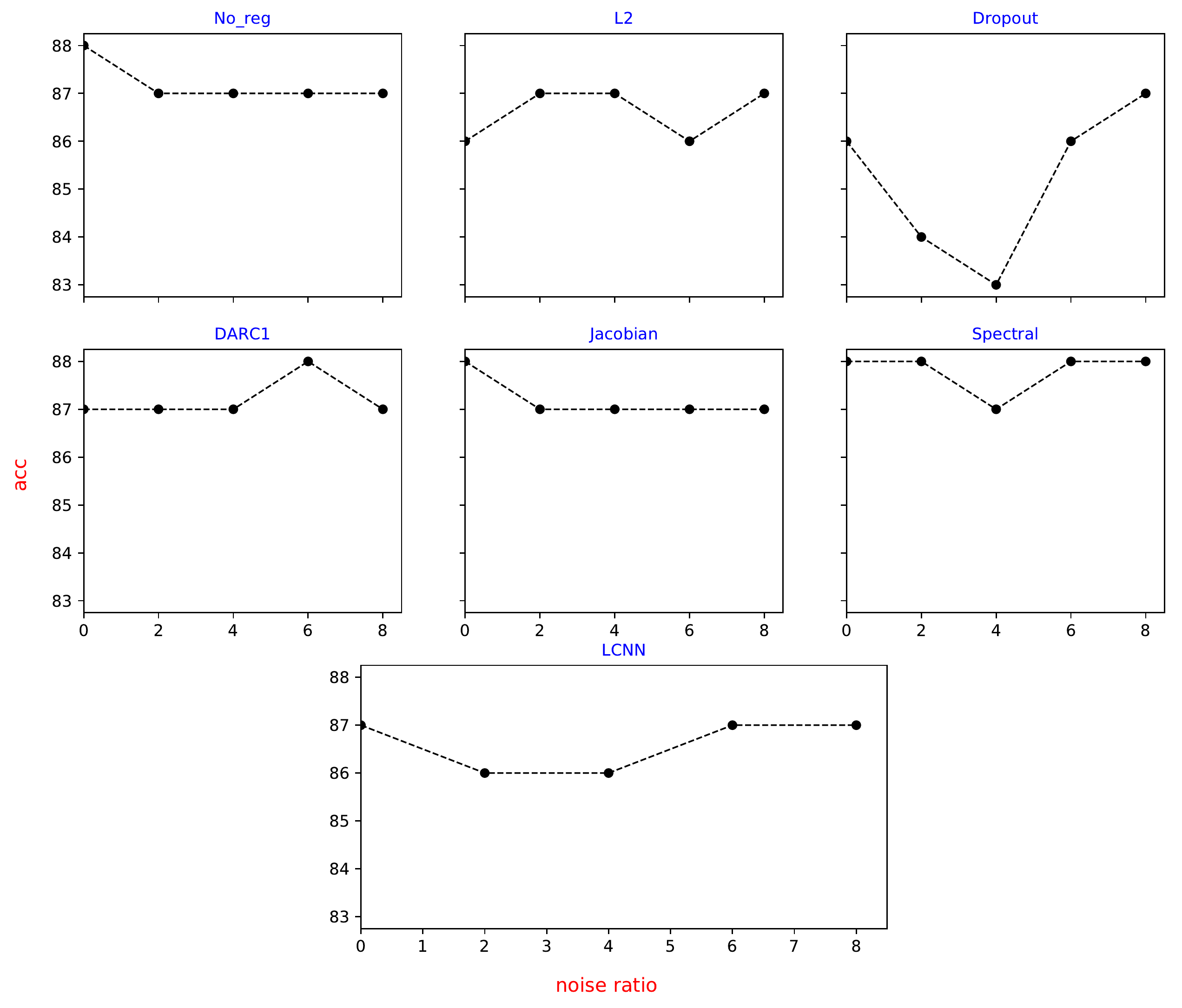}
		\caption{}
		\label{fig:acc 14}	
	\end{subfigure}\qquad
	\begin{subfigure}[hbtp]{0.5\textwidth}
		\includegraphics[width=\textwidth]{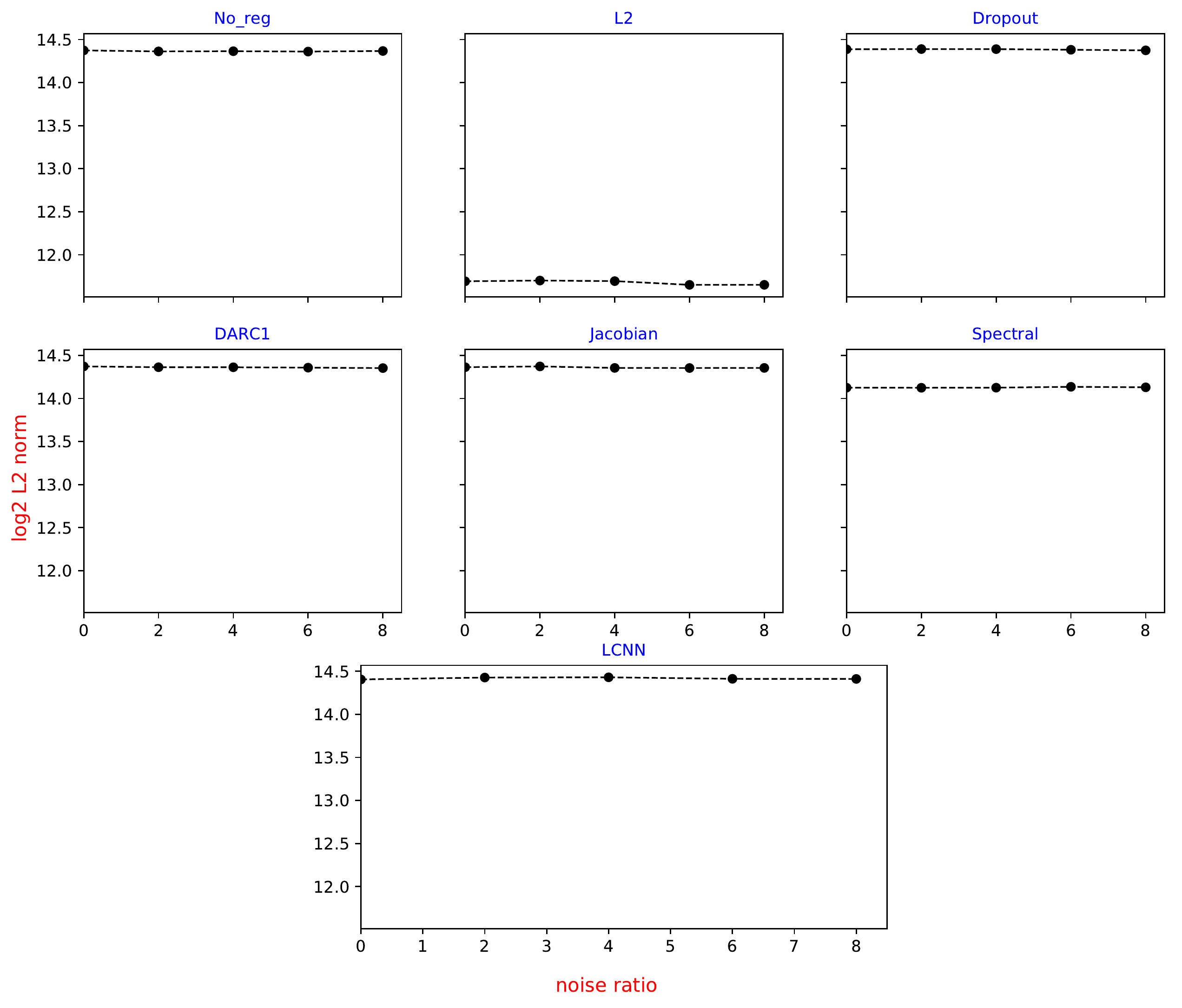}
		\caption{}
		\label{fig:log l2 14}	
	\end{subfigure}
	
	\begin{subfigure}[hbtp]{0.5\textwidth}
		\includegraphics[width=\textwidth]{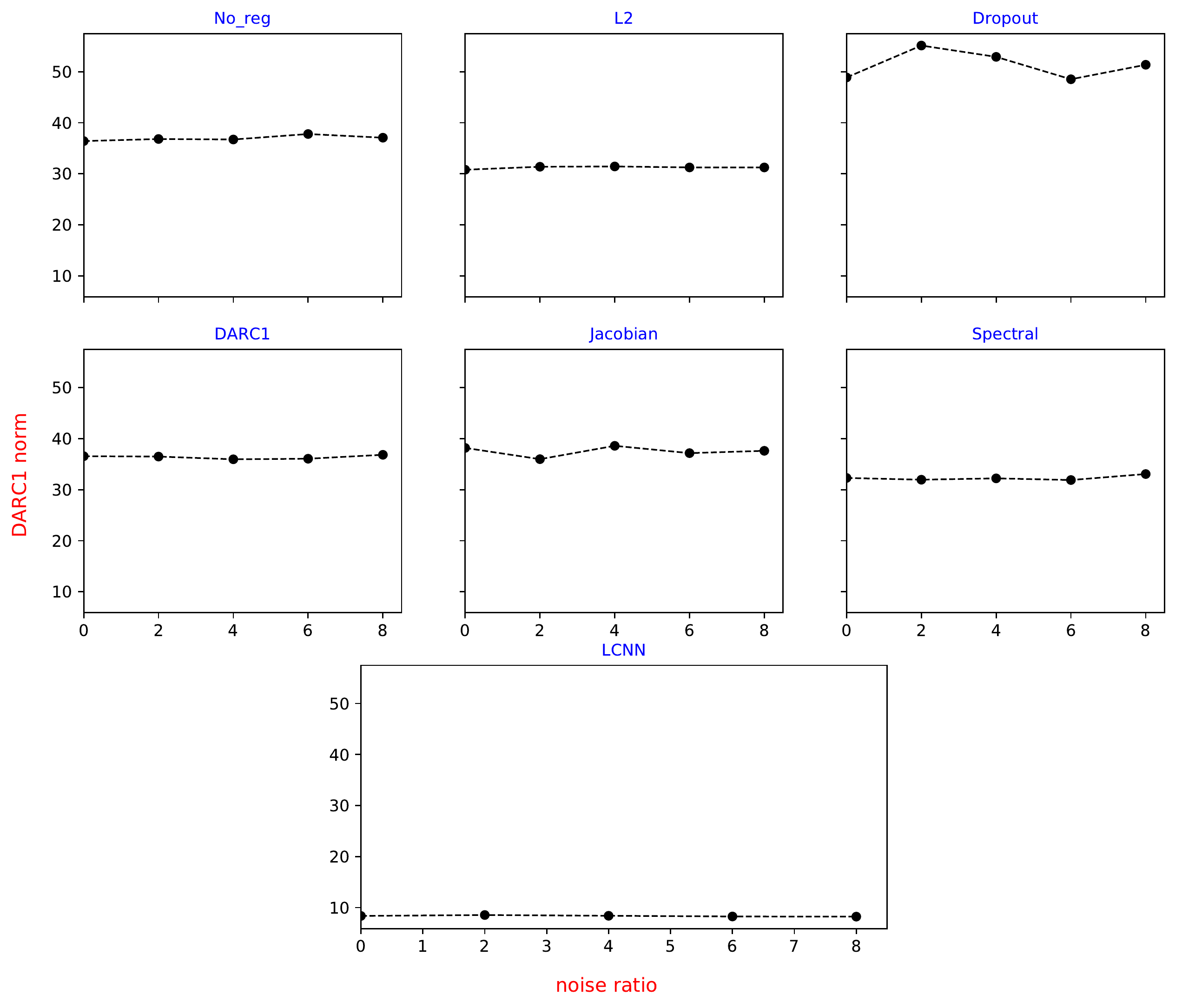}
		\caption{}
		\label{fig:darc1 14}	
	\end{subfigure}\qquad
	\begin{subfigure}[hbtp]{0.5\textwidth}
		\includegraphics[width=\textwidth]{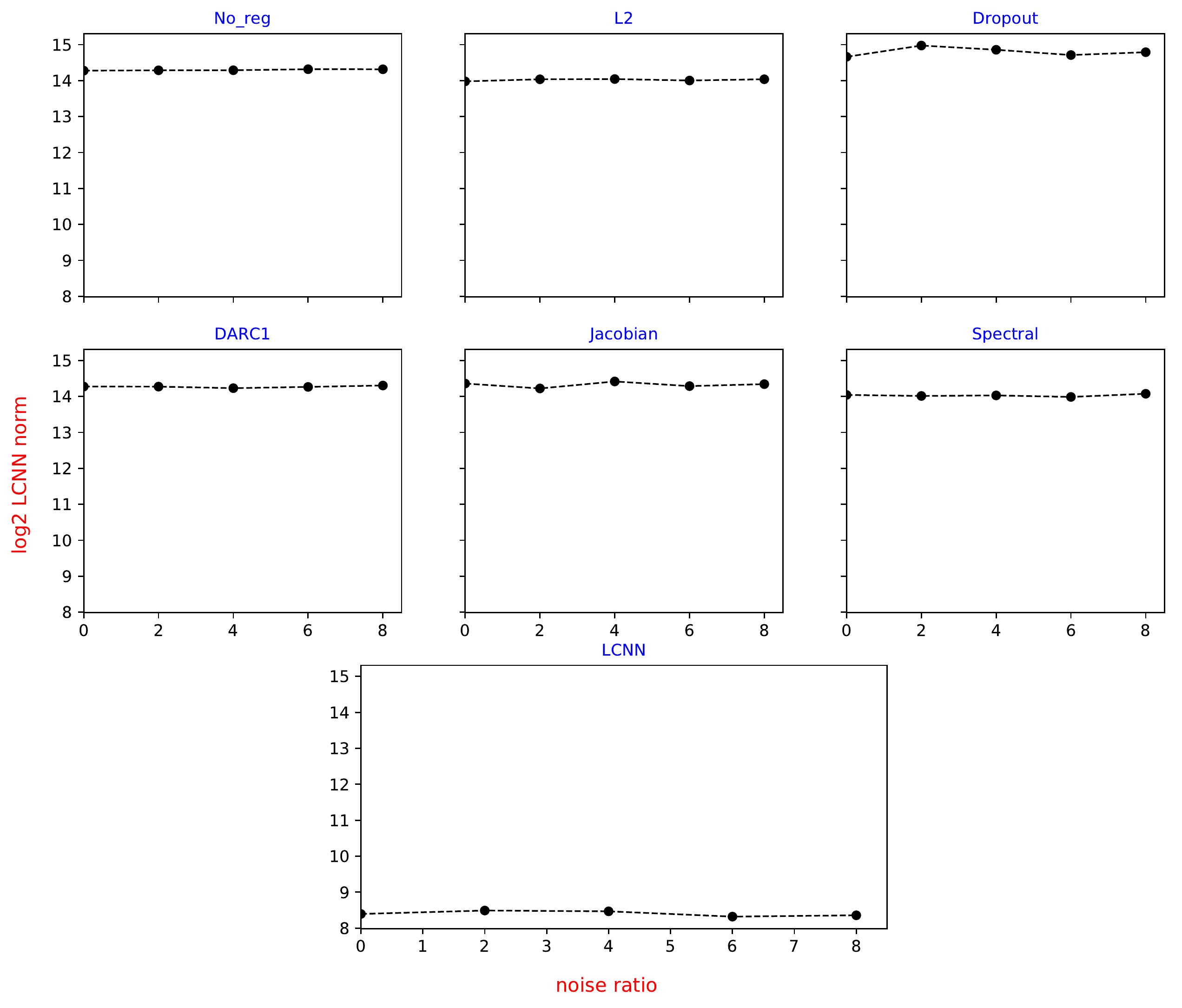}
		\caption{}
		\label{fig:log lcnn 14}	
	\end{subfigure}
	
	\begin{subfigure}[hbtp]{0.5\textwidth}
		\includegraphics[width=\textwidth]{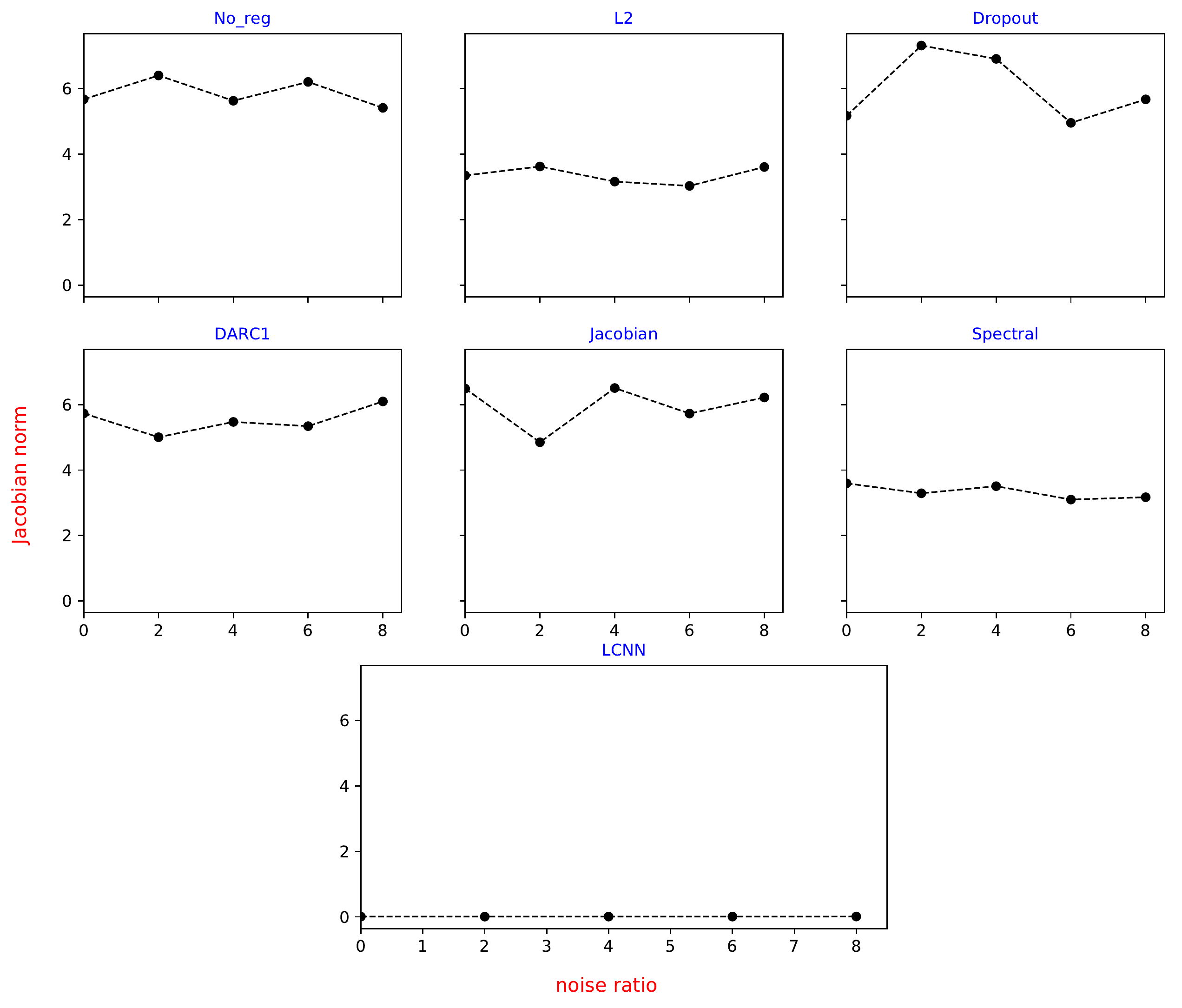}
		\caption{}
		\label{fig:jacobian 14}	
	\end{subfigure}\qquad
	\begin{subfigure}[hbtp]{0.5\textwidth}
		\includegraphics[width=\textwidth]{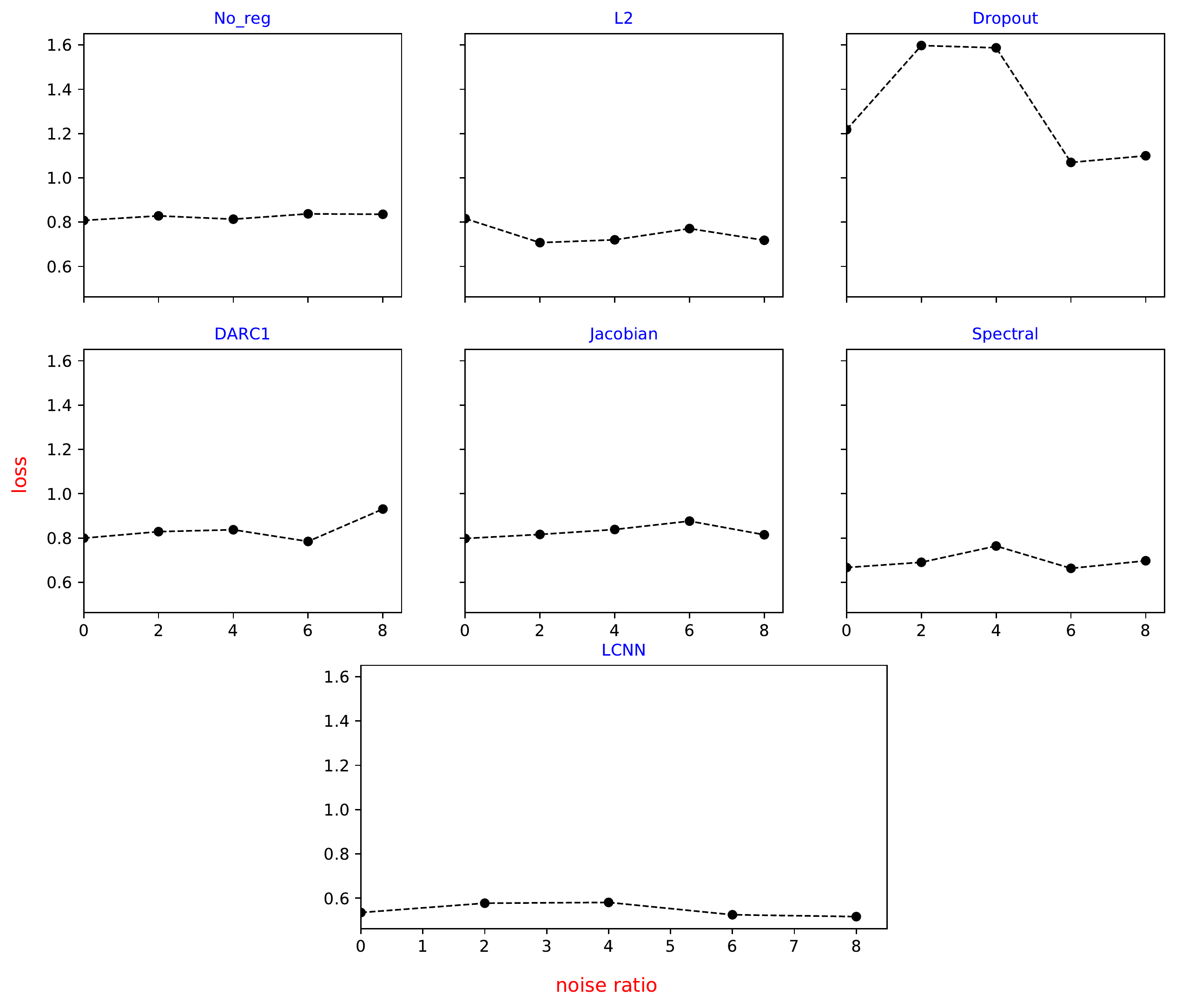}
		\caption{}
		\label{fig:loss 14}	
	\end{subfigure}
	\caption{Comparison of \subref{fig:acc 14} test set accuracies, \subref{fig:log l2 14} $L_2$ norm of weights, \subref{fig:darc1 14} mean DARC1 regularizer, \subref{fig:log lcnn 14} mean LCNN regularizer, \subref{fig:jacobian 14} mean Jacobian regularizer, \subref{fig:loss 14} mean cross entropy loss for shallow network across datasets}
\label{fig:comparison cifar}
\end{figure}

Figure \ref{fig:acc 14} shows the accuracies obtained as the noise ratio  increases for various regularization techniques. We observe that dropout results in a non-robust graph, where the accuracy drops sharply as noise increases earlier and then increases with noise. Spectral normalization results in the highest set of accuracies and is not affected by noise. For other methods such there is no appreciable drops in accuracies as the noise increases, thus validating the hypothesis that deep neural nets are robust to input noise.

Figure \ref{fig:log l2 14} shows the log2 $L_2$ norm. We see that for $L_2$ regularization, the $L_2$ norm is the smallest, followed by spectral norm. The $L_2$ norm in case of $L_2$ is orders of magnitude smaller than the rest, despite having comparable accuracies. This is indicative of the fact that deeper architectures have high model complexities which be controlled using $L_2$ norm.

Figures \ref{fig:darc1 14} and \ref{fig:log lcnn 14} shows the DARC1 and LCNN norm respectively. It can be seen that LCNN regularizer results in smallest respective norms, followed by Spectral norm and $L_2$ norm. Dropout results in the largest DARC1 and LCNN norms.

Figure \ref{fig:jacobian 14} shows the Jacobian norm for various regularizers. We observe that for LCNN norm the Jacobian norm is the smallest. This shows that the error signal is not able to propagate to the initial layers. This may be due to high value of LCNN hyperparameter. Among others, Spectral norm shows the smallest values for the Jacobian norm and also the smallest variation. This also corresponds to the robustness in accuracies as the noise increases. Dropout shows the highest variation in Jacobian norm which corresponds to the large variation in accuracies as the noise increases. It is also observed that Jacobian norm increases with noise, which is indicative of increasing uncertainity in data.

Figure \ref{fig:loss 14} shows the cross entropy loss variation with noise. The loss is inversely related to the accuracy (fig. \ref{fig:acc 14}). The loss is high for network with only Dropout as regularizer. The loss is minimal for LCNN norm, followed by Spectral normalization and $L_2$ norm.

\section{Discussion}
\label{sec:discussion}
From the results we can obtain the following insights:
\begin{enumerate}
	\item The model complexity can be fairly represented using $L_2$ norm, DARC1 norm and LCNN norm.
	\item $L_2$ regularization and DARC1 regularization performs well in controlling the model complexity of the network. Their smaller values are indicative of a less complex hypothesis class learnt in the wake of higher regularization. 
	\item As the noise increases for shallow networks, the $L_2$ norm, DARC1 norm increases at first and then we observe a decrease in the norms indicative of the simpler hypothesis learn when the validation set is noise free.
	\item For shallow network, Jacobian norm performs well in the presence of noise. For deeper architectures, network itself performs noise reduction and thus Jacobian regularization has minimal affect. Deeper networks are robust to input noise as is elucidated in Arora et al., \cite{arora2018stronger}.
	\item Spectral normalization performs the best in case of noisy inputs closely followed by DARC1 regularizer which is derived from a distribution dependent bound.
	\item The test set loss is a good indicator of the test set accuracy.
	\item Dropout alone does not perform well in case of noisy inputs neither does it control the model complexity given in terms of $L_2$ norm, DARC1 norm or LCNN norm.
	\item Adding a input noise can result in better generalization properties for deeper architectures, however this effect is less pronounced for shallow architectures.
\end{enumerate}

\section{Conclusion}
\label{sec:conclusion}
In this paper we presented a study on the effect of regularization of model complexity and generalization as the noise in the inputs $x \in \mathcal{X}$ increases. We find multiple notions of model complexities in the literature ranging from distribution independent VC dimension bounds for neural networks \cite{sontag1998vc,sharma2018radius} to distribution dependent Rademacher complexity bounds \cite{bartlett2002rademacher,neyshabur2015norm,neyshabur2017exploring,bartlett2017spectrally}. All these bounds are in terms of product or sum of norm of weights of the network. We used sum of $L_2$ norm of the weights as proxy for model complexity. This directly translates to adding a weight decay regularizer that penalizes larger weights. Recently proposed DARC1 or LCNN regularizer are a step in generating a distribution dependent bounds for neural networks. We see the both $L_2$ and DARC1 controls the model complexity.

The experiments clearly demonstrate that the $L_2$ norm is best suited for controlling model complexity as well as for generating solutions with high accuracies. For deeper networks Dropout alone does not result in higher accuracies or control the model complexity in terms of the norm of weights. We also see that the newly proposed distribution dependent DARC1 and LCNN regularization performs equally well as $L_2$ regularization. Spectral normalization results in one of the best results for deeper architectures, due to stable gradient propagation.
Our experiments also show that deep neural networks are more generally more robust to varying degrees of Gaussian input noise than shallow architectures and can therefore effectively model any number of such distributions under strong priors. In future, we shall explore novel regularization schemes like shakedrop \cite{yamada2018shakedrop} and shake-shake regularization \cite{gastaldi2017shake}. We have refrained from discussing recurrent architectures like Long short term memory (LSTM) network \cite{gers1999learning,greff2017lstm}, which shall be the focus of our future research.

\bibliographystyle{IEEEtran}
\bibliography{bibliography}
\end{document}